\documentclass[10pt, conference, compsocconf]{IEEEtran}
\usepackage[switch]{lineno}

\usepackage{cite}
\usepackage{amsmath,amssymb,amsfonts}
\usepackage{algorithmic}
\usepackage{graphicx}
\usepackage{textcomp}
\usepackage{xcolor}

\usepackage{balance}
\usepackage{booktabs} 
\usepackage{multirow}
\usepackage{graphicx}
\usepackage{subfigure}
\usepackage{setspace}
\usepackage{amssymb}
\usepackage{array}
\usepackage{amsmath}
\usepackage{times}  
\usepackage{helvet} 
\usepackage{courier} 
\usepackage[numbers]{natbib}
\usepackage{enumitem}
\usepackage{kantlipsum}
\usepackage{url}

\begin{document}

\title{Building Autocorrelation-Aware Representations for Fine-Scale Spatiotemporal Prediction}

\makeatletter
\newcommand{\linebreakand}{%
  \end{@IEEEauthorhalign}
  \hfill\mbox{}\par
  \mbox{}\hfill\begin{@IEEEauthorhalign}
}
\makeatother

\author{ 
\IEEEauthorblockN{Yijun Lin}
\IEEEauthorblockA{Department of Computer Science \\
University of Southern California\\
yijunlin@usc.edu}
\and
\IEEEauthorblockN{Yao-Yi Chiang}
\IEEEauthorblockA{Spatial Sciences Institute \\
University of Southern California\\
yaoyic@usc.edu}
\and
\IEEEauthorblockN{Meredith Franklin}
\IEEEauthorblockA{Department of Preventive Medicine \\
University of Southern California\\
meredith.franklin@usc.edu}
\linebreakand % <------------- \and with a line-break
\IEEEauthorblockN{Sandrah P. Eckel}
\IEEEauthorblockA{Department of Preventive Medicine \\
University of Southern California\\
eckel@usc.edu}
\and
\IEEEauthorblockN{Jos\'e Luis Ambite}
\IEEEauthorblockA{Information Sciences Institute \\
University of Southern California\\
ambite@isi.edu}
}

\maketitle
\newcommand{\todo}[1]{{\color{blue}[TODO: #1]}}
\newcommand{\KLD}{D_{\mathrm{KL}}\left( P||Q\right)}

\begin{abstract}
Many scientific prediction problems have spatiotemporal data- and modeling-related challenges in handling complex variations in space and time using only sparse and unevenly distributed observations. This paper presents a novel deep learning architecture, Deep learning predictions for LocATion-dependent Time-sEries data (DeepLATTE), that explicitly incorporates theories of spatial statistics into neural networks to addresses these challenges. In addition to a feature selection module and a spatiotemporal learning module, DeepLATTE contains an autocorrelation-guided semi-supervised learning strategy to enforce both local autocorrelation patterns and global autocorrelation trends of the predictions in the learned spatiotemporal embedding space to be consistent with the observed data, overcoming the limitation of sparse and unevenly distributed observations. During the training process, both supervised and semi-supervised losses guide the updates of the entire network to: 1) prevent overfitting, 2) refine feature selection, 3) learn useful spatiotemporal representations, and 4) improve overall prediction. 
We conduct a demonstration of DeepLATTE using publicly available data for an important public health topic, air quality prediction, in a well-studied, complex physical environment - Los Angeles. The experiment demonstrates that the proposed approach provides accurate fine-spatial-scale air quality predictions and reveals the critical environmental factors affecting the results. 
\end{abstract}


\section{Introduction} 
% \linenumbers
% intuition of the paper
Fine-scale spatiotemporal prediction is an important scientific problem applicable to diverse phenomena, such as air quality, ambient noise, traffic conditions, and meteorology.\footnote{Here, the prediction problem refers to estimating measurements for unobserved locations at current or past times (in contrast to forecasting future measurements).}  One of the primary motivations for spatiotemporal prediction is that the observed data are only available at a few unevenly distributed measurement locations (e.g., ground-based sensors)~\cite{jiang2018survey, karroum2020review}.
As an example, there is only a handful of Federal monitoring sites and a few hundred low-cost sensors reporting air quality in Los Angeles, an area that covers nearly 5,000 square miles and 15 million people. In many epidemiological studies, the exposure assessment usually needs the air quality at a fine spatial scale for analyzing the associations between inhalable particles\footnote{ For example, fine particles with an aerodynamic diameter smaller than 2.5$\mu$m, i.e., PM$_{2.5}$.} and a variety of health outcomes including cardio-respiratory diseases~\cite{feng2016health}. 
However, sparse ground-based monitoring networks do not provide the spatial resolution required to characterize exposures where people spend their time: home, work, and school~\cite{lin2018exploiting,lin2017mining}.

% some existing methods in general
Approaches for predicting spatiotemporal phenomena generally fall into the categories of expert-based and data-driven methods. Expert-based air quality prediction methods, such as atmospheric dispersion models, utilize mathematical equations to simulate how pollutants travel in the physical environment. They require significant expert knowledge and computational resources (e.g., modeling chemical transformations in the atmosphere~\cite{karroum2020review, leelHossy2014dispersion}). In contrast, data-driven methods aim to characterize spatiotemporal patterns in the available measurement data and leverage the patterns for prediction. One common data-driven approach is spatial interpolation, such as Inverse Distance Weighting (IDW) and Ordinary Kriging (OK), which utilize spatial distance-based relationships (in either non-geostatistical or geostatistical manner) between observed and target locations to make predictions where there are no observations~\cite{li2011review}. Spatial interpolation methods do not consider external environmental characteristics, such as meteorology or land use. Hence, they have limited ability to produce reliable fine-scale predictions (e.g., would have similar prediction values across a large area)~\cite{lin2017mining}. 

Advanced data-driven methods take various contextual data about the environment into account to improve the robustness of the prediction model and outcomes. The main challenge is how to automatically discover meaningful spatial and temporal relationships between rich contextual data and sparse measurement data in an area. For example, land-use regression (LUR) models generate explainable air quality predictions by including expert-selected land-use predictors (e.g., traffic indicators, industrial facilities, and population density)~\cite{hoek2008review}, which can vary from one geographic region to another~\cite{zheng2013u}. Machine learning models, such as neural networks and random forests, can alleviate the requirement of expert-selected predictors in the contextual data by algorithmically choosing the most useful features for prediction tasks (e.g.,~\cite{zheng2013u,hsieh2015inferring,qi2018deep, lin2017mining}). However, existing methods only take neighborhood-level spatial or temporal dependencies into consideration (e.g.,~\cite{qi2018deep}), which could suffer from overfitting when only a limited number of measurement locations are available, or the measurement locations are sparse and not spatially representative of the target region.

This paper proposes a novel data-driven approach, named DeepLATTE (Deep learning prediction for LocATion-dependent Time-sEries data), that explicitly incorporates theories of spatial statistics into neural networks for fine-spatial-scale prediction of geographically-based spatiotemporal phenomena. The goal\footnote{We demonstrate DeepLATTE within the air quality domain, but the overall approach does not assume a specific application domain.}  is to estimate air quality values at locations that do not have sensors over a fine spatial grid by using the observed air quality measurements from a network of low-cost sensors as well as the contextual data describing the environmental characteristics. First, DeepLATTE automatically generates a condensed feature embedding of the contextual data for both labeled (measurement) and unlabeled (target) locations using a standard sparse layer and an auto-encoder. These feature embeddings then go through a Convolutional Long Short-Term Memory (ConvLSTM) module that incorporates the contextual information in local spatial and temporal neighborhoods to generate spatiotemporal embeddings for all locations. Next, DeepLATTE employs a novel autocorrelation-guided semi-supervised learning strategy that 1) enforces the spatial and temporal neighboring embeddings to be similar (i.e., the learned environmental characteristics should change gradually or slightly over a small range of space and time) and 2) enforces the distribution of the predicted values and the distribution of the labels in the embedding space to be similar. 
The first step helps propagate useful information within a local region in space and time to refine the spatiotemporal embeddings of all locations. The second process is inspired by Kriging using a semivariogram, which ensures the predictions show similar graduated autocorrelation patterns in the learned spatiotemporal embedding space as in the labeled data. This semi-supervised learning strategy prevents the networks from focusing on only the labeled locations and hence can fully utilize the contextual data for each location and time to produce accurate prediction results.

In sum, the main contribution of this paper is a general network architecture that exploits both local autocorrelation patterns and global autocorrelation trends of the predictions and the labeled data in the learned spatiotemporal embedding space for accurate fine-scale prediction task using limited labeled data. Also, we show an end-to-end approach using this network architecture for fine-scale air quality prediction. 

\section{DeepLATTE Methodology}
This section presents the proposed approach, DeepLATTE, for fine-spatial-scale prediction of spatiotemporal data using air quality prediction as an example. The inputs include the contextual data and the available measurements represented in a grid structure (tensors). The contextual data are denoted as $\boldsymbol{X} \in \mathbb{R}^{H \times W \times P}$, where $P$ is the number of input features, and $H$ and $W$ are the height and width of the grid, respectively. The measurements are denoted as $\boldsymbol{Y}  \in \mathbb{R}^{H \times W}$, which contain many missing values. The prediction results are denoted as $\boldsymbol{\hat{Y}} \in \mathbb{R}^{H \times W}$. Let $\boldsymbol{X}^{(t)}$ represent the input signal at time $t$, $T'$ is the number of previous hours (i.e., from $t-T'+1$ to $t$). DeepLATTE aims to learn a function $h$ that maps $T'$ historical input signals to the output at time $t$:

\begin{displaymath}
[\boldsymbol{X}^{(t-T'+1)}, \cdots, \boldsymbol{X}^{(t)}]\xrightarrow{\emph{h}}[\boldsymbol{\hat{Y}}^{(t)}]
\end{displaymath}

Figure~\ref{figure: whole} shows the overall architecture of DeepLATTE. 
After constructing the input grid data~$\boldsymbol{X}$ and labeled data~$\boldsymbol{Y}$ (Section~\ref{sec:ggd}), DeepLATTE uses a sparse layer (Section~\ref{sec:fs}) and an encoder-decoder module (Section~\ref{sec:life}) to generate a condensed feature embedding for the selected predictors. DeepLATTE then leverages multiple ConvLSTM layers with varying kernel sizes to learn a spatiotemporal representation of embeddings and enforces spatially and temporally nearby embeddings to be similar (Section~\ref{sec:lrsi}). Afterwards, DeepLATTE constrains the global autocorrelation trends of the predictions to be close to those of the labeled data in the learned space of the spatiotemporal embeddings in a semi-supervised way (Section~\ref{sec:ssl}). This approach assumes that natural and built environments (i.e., contextual data) contribute to the target spatiotemporal phenomenon (e.g., air quality), and the predictions are highly correlated when their learned representations of the environments are similar. The approach does not assume a specific application domain.

\begin{figure*}[htbp]
  \centering
  \includegraphics[width=0.87\linewidth]{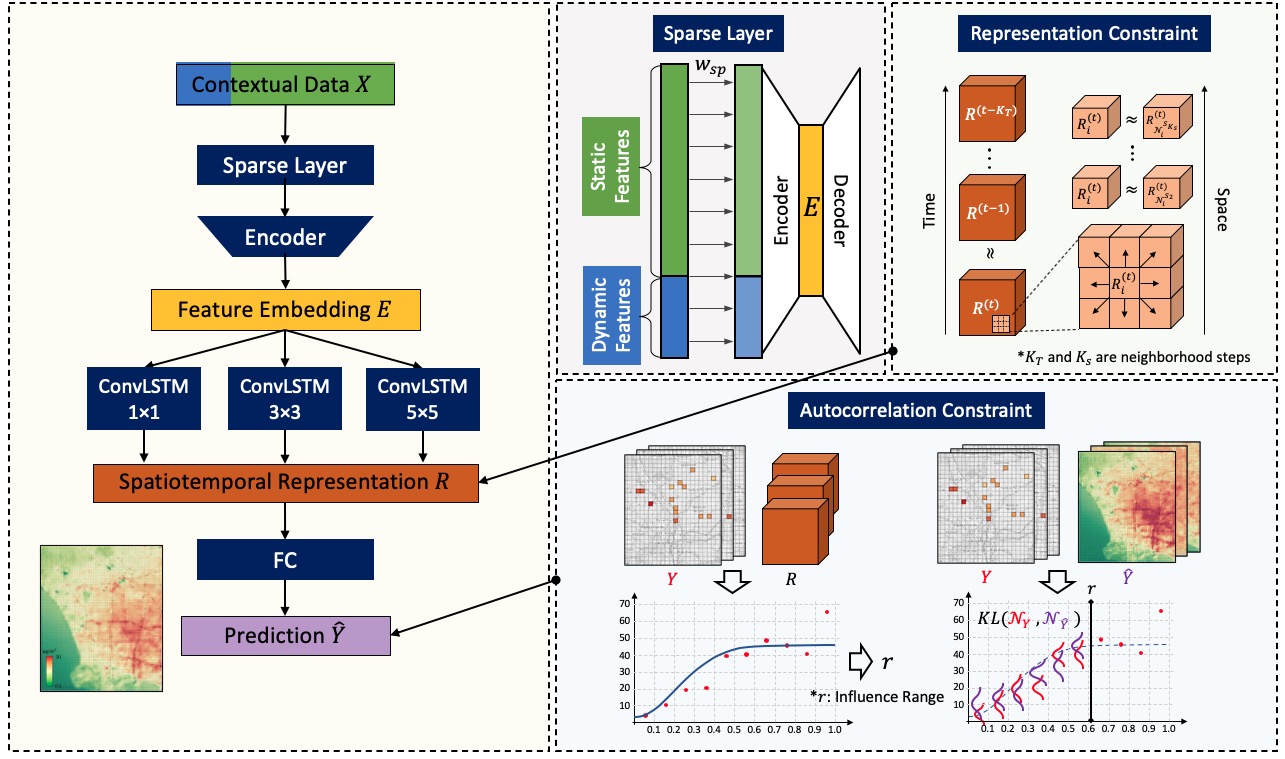}
  \vspace{-.1in}
  \caption{DeepLATTE network architecture}
  \label{figure: whole}  
  \vspace{-0.13in}
\end{figure*}

\subsection{Generating Grid Data}\label{sec:ggd}
The raw data (i.e., labeled air quality data from sensors and contextual data) can have varying spatial and temporal resolutions. To unify and aggregate the input data, DeepLATTE divides the target area into disjointed cells and transforms the raw data into a grid representation.
Each cell contains a set of features, which represent an aggregation of the contextual data. We denote $\boldsymbol{X}_{i} \in \mathbb{R}^P$ as the input vector of cell $i$ on the grid. Let $\boldsymbol{D}_{i}\in \mathbb{R}^{P_d}$ represent the dynamic (time-varying) features (e.g., weather) and $\boldsymbol{S}_{i}\in \mathbb{R}^{P_s}$ represent the static (time-invariant) features (e.g., built environment). $P_d$ and $P_s$ are the number of dynamic features and static features, respectively. If the 
measurements of a feature are sparser than the grid resolution but uniformly distributed in space (e.g., weather data), DeepLATTE spatially up-scales the feature with cubic interpolation; otherwise, DeepLATTE either directly adopts the measurements for a cell (e.g., coordinate) or aggregates the measurements within the cell (e.g., the area of parks) to construct the feature vector. Figure~\ref{figure: roads} shows an example of one feature component, primary roads, in the spatial grid covering Los Angeles. The cell value is the summation of the lengths of individual primary roads within the cell.

This process assumes that the observations in a cell are uniform (i.e., one value per cell). The system maps the available sensors to the corresponding grid cells to generate labels. If one cell contains multiple sensors, the cell takes the average of their sensor readings. Figure~\ref{figure: pm25} shows an example grid map of PM$_{2.5}$ observations. The colored cells are the locations with ground truth measurements (i.e., labeled locations), while the uncolored cells are the target (unlabeled) locations for prediction.
% The goal is to predict the PM$_{2.5}$ concentrations for every cell in the PM$_{2.5}$ grid map. 

% HACK: moved here for float placement
\begin{figure}[htbp]
\vspace{-0.05in}
\centering
    \subfigure[Grid primary roads]{
        \begin{minipage}[t]{0.46\columnwidth}
            \centering
            \includegraphics[width=\linewidth]{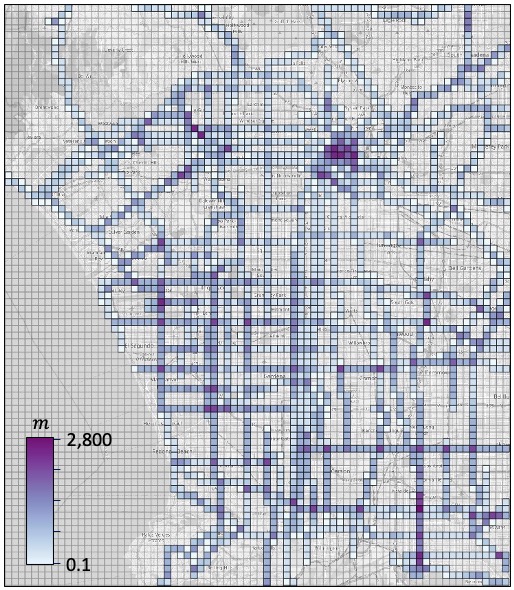}
            \vspace{-0.1in}
            \label{figure: roads}
        \end{minipage}
    }
    \subfigure[Grid PM$_{2.5}$ concentrations]{
    \begin{minipage}[t]{0.46\linewidth}
        \centering
        \includegraphics[width=\linewidth]{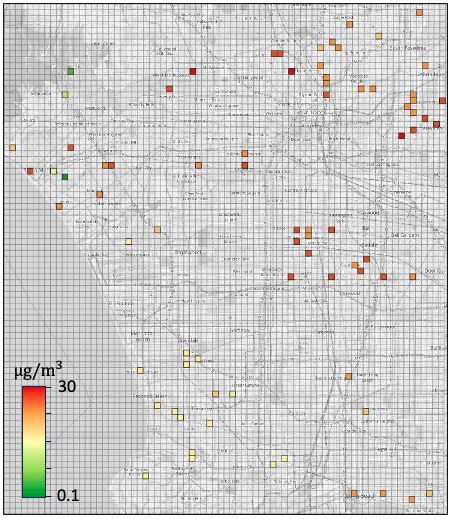}
        \vspace{-0.1in}
        \label{figure: pm25}  
    \end{minipage}
    }
\vspace{-0.05in}
\caption{An example grid map with 500m$\times$500m cells covering a 50km$\times$40km area in Los Angeles. (a) The aggregated length of primary roads in the cells. (b) PM$_{2.5}$ concentrations after mapping sensor data to the grid. Missing values and zeros are not colored.}
\vspace{-0.12in}
%\caption{An example map with cell size of 500m$\times$500m covering 30km$\times$30km area in Los Angeles. (a) shows the temperature variations after the interpolation. (b) shows the accumulative length of primary roads on the same grid map. The darker the color, the more primary roads in the cell. The uncolored cells do not contain value inside.}
\end{figure}

\subsection{Feature Selection}\label{sec:fs}
DeepLATTE first uses a sparse layer with $L_1$ regularization (Figure~\ref{figure: whole}, ``Sparse Layer'') with the purpose of (1) removing irrelevant input features and (2) balancing the contributions of dynamic and static features (e.g., in our air quality case, there many more static features than dynamic ones). 

The sparse layer is a linear layer containing the same number of nodes as the input, and there exists only a single connection between the corresponding nodes. Let $W_{sp}$ be the weight matrix of the sparse layer, which is a diagonal matrix (zero off-diagonal weights). We add $L_1$ regularization as the sparse constraint to this layer to force the sum of the absolute values of the parameters to be small. Thus, the $L_1$ penalty term can cause many weights to be close to zero. If the weight is a tiny value (less than a predefined threshold), DeepLATTE sets the output of the sparse layer of the corresponding feature to zero to achieve the purpose of feature selection. We denote the output of the sparse layer as $\boldsymbol{X}_{sp} \in \mathbb{R}^{H \times W \times P}$. The dimension of $\boldsymbol{X}_{sp}$ is the same as $\boldsymbol{X}$ with irrelevant features being zeros. The cost function of $L_1$ regularization is as follows:
\begin{equation}
    \mathcal{L}_{sp}={\sum_{w \in W_{sp}}{|w|}}
\label{eq: sp}
\end{equation}

\subsection{Learning Condensed Feature Embeddings}\label{sec:life} 
% \vspace{-0.03in}
DeepLATTE pre-trains an auto-encoder to learn a condensed feature embedding $\boldsymbol{E}_{i}$ for a cell $i$ from the selected dynamic $\boldsymbol{D}_{i}$ and static $\boldsymbol{S}_{i}$ features (Figure~\ref{figure: whole} ``Encoder'' and ``Decoder''). For example, suppose primary road is an indicator of traffic volume, and its impact can vary with time, weather conditions, and location. It is necessary to learn a latent embedding to condense these two types of features instead of directly feeding them original features vector to the network. The reconstruction loss for the auto-encoder is:

\begin{equation}
    \mathcal{L}_{ae}={\mathcal{\bar{L}}(\boldsymbol{X}_{sp}, \boldsymbol{\hat{X}}_{sp})}
\label{eq: ae}
\end{equation}

\noindent where $\boldsymbol{X}_{sp}$ is the output of the sparse layer (the encoder input), $\boldsymbol{\hat{X}}_{sp}$ is the reconstructed vector (the decoder output), and $\mathcal{\bar{L}}$~is the loss function using the mean square error. During the end-to-end training after pre-training, the network can refine the condensed feature embedding $\boldsymbol{E}$
% to capture the interactive effects between components in $\boldsymbol{X}_{sp}$ 
by updating the encoder weights guided from the remainder of the network modules. % and the labeled data. 
\vspace{-0.05in}

\subsection{Learning Spatiotemporal Feature Embeddings}\label{sec:lrsi}
Current air quality level at a location is highly correlated with the environmental characteristics at the location and its neighbors at present and previous time points. For example, suppose a South-West power plant emits pollution at time $T$, the northeast cells can be significantly polluted at $T+1$ with a northeast wind (Figure \ref{figure: spatial_temporal_influence_all}).  

We define the spatiotemporal embedding of a cell $i$ at time $t$ as the joint effects of the air quality-related factors in the condensed feature embeddings from neighboring space and time, denoted as $\boldsymbol{R}^{(t)}_{i}$. To jointly model the spatiotemporal impacts on air quality, DeepLATTE first leverages the ConvLSTM operation~\citep{xingjian2015convolutional} to learn useful information from the combination of the condensed embeddings from spatial neighbors and the previous hidden memory. For example, in Figure~\ref{figure: spatial_temporal_influence}, when generating the spatiotemporal embedding of the center cell (red box) using one-step neighbors only (in the purple dotted box), DeepLATTE learns the interactive effects from the northeast green areas, the northwest residential areas, the South-East industrial areas, and the East commercial areas. To account for the influence of environmental characteristics on air quality from varying distances, DeepLATTE employs multiple ConvLSTM layers with various kernel sizes to learn the impacts from the neighbors within increasing spatial distances. The outputs of these ConvLSTM layers with varying kernel sizes are then concatenated to form an initial spatiotemporal embedding for each cell.

Due to the limited number of sparse labeled data, the initial spatiotemporal embeddings would be only learned from the labeled data. To overcome this difficulty, DeepLATTE adds a representation constraint in the learning process by enforcing the spatially and temporally nearby spatiotemporal embeddings to be similar with the assumption that the embeddings would not change significantly within a local spatial and temporal neighborhood. This constraint guides the network to produce a graduated, continuous spatiotemporal representation from all input data instead of focusing on the labeled locations. The cost function is as follows: 

\begin{equation}
\begin{aligned}
\mathcal{L}_{stc} = \lambda_1 \times \sum_{i=1}^{N} \sum_{k=1}^{K_S} \sum_{j \in {\mathcal{N}^{(S_k)}_i}} \frac{1}{k}\mathcal{\bar{L}}(\boldsymbol{R}_i, \boldsymbol{R}_j) \\
+ \lambda_2 \times \sum_{i=1}^{N} \sum_{k=1}^{K_T} \sum_{j \in {\mathcal{N}^{(T_k)}_i}} \frac{1}{k} \mathcal{\bar{L}}(\boldsymbol{R}_i, \boldsymbol{R}_j)
\end{aligned}
\label{eq: stc}
\end{equation}

\noindent where $\boldsymbol{R}_i$ is the spatiotemporal embedding for cell $i$ at a certain time. $N$ is the total number of spatiotemporal embeddings (the number of target locations multiplied by the number of time points for prediction). $K_S$ and $K_T$ are the number of neighborhoods to consider in space and time, respectively, where $k=1$ implies the immediate, one-step neighborhood, and $k=2$ is two-step neighborhood.
${\mathcal{N}^{(S_k)}_i}$ is the neighbors of the cell $i$ within $k$ spatial step. Similarly, ${\mathcal{N}^{(T_k)}_i}$ represents the temporal neighbors. $\mathcal{\bar{L}}$ is the loss function of mean square error. This cost function ensures that nearby spatiotemporal embeddings in space and time are similar, and the similarity gradually decreases when the distance increases. Finally, the refined spatiotemporal embeddings are fed to fully connected layers to infer air quality prediction for each cell. 
%In the experiments, we simply set $K_T$ and $K_S$ to 1 and hyper-parameters $\lambda_1=\lambda_2$.

% HACK: moved here for float placement
\begin{figure}[htbp]
\vspace{-0.08in}
\centering
    \subfigure[Time $T$]{
    \begin{minipage}[t]{0.29\columnwidth}
        \centering
        \includegraphics[width=\linewidth]{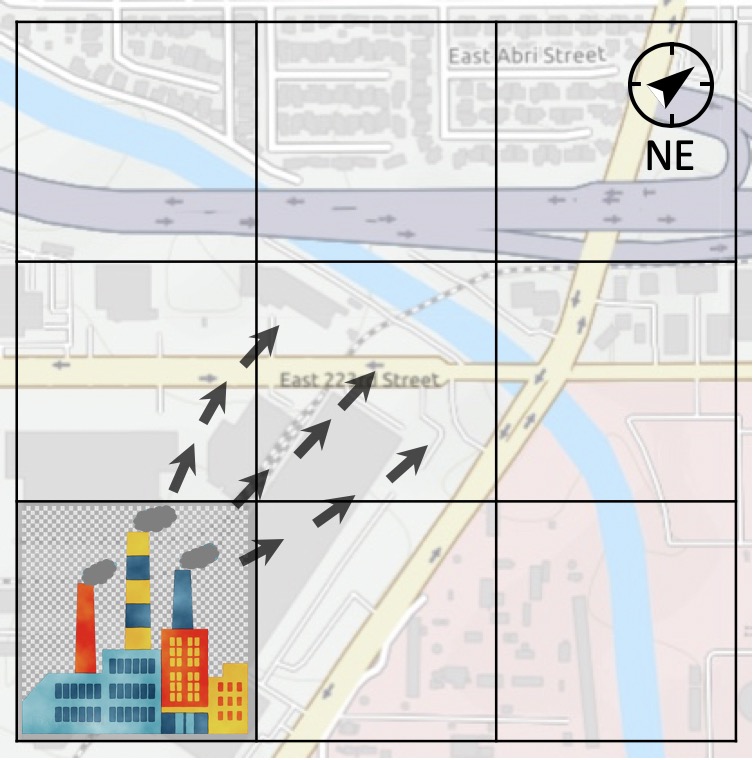}
        \vspace{-0.1in}
        \label{figure: T}  
    \end{minipage}
    }
    \subfigure[Time $T+1$]{
        \begin{minipage}[t]{0.29\columnwidth}
            \centering
            \includegraphics[width=\columnwidth]{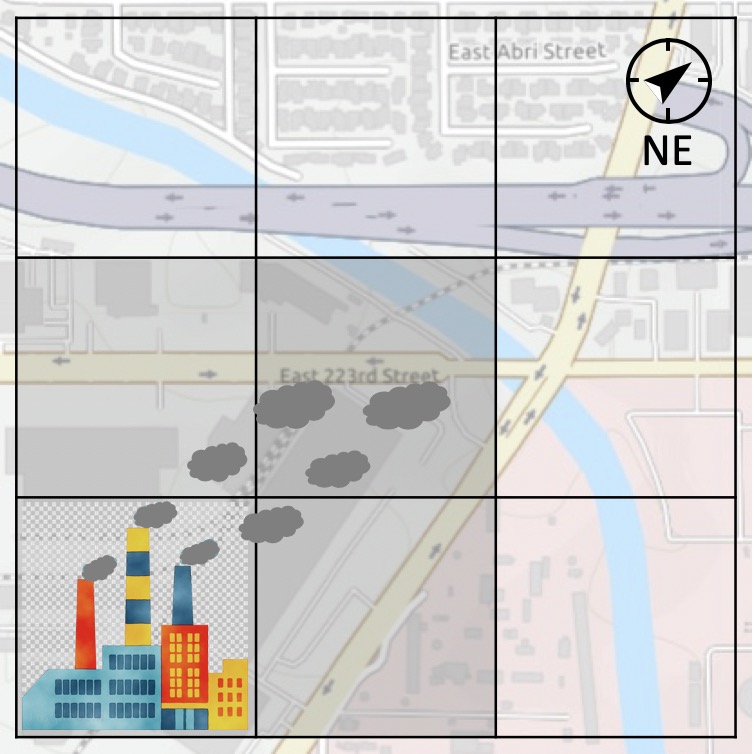}
            \vspace{-0.1in}            
            \label{figure: T+1}
        \end{minipage}
    }
    \subfigure[Spatial Effects]{
        \begin{minipage}[t]{0.29\columnwidth}
            \centering
            \includegraphics[width=\columnwidth]{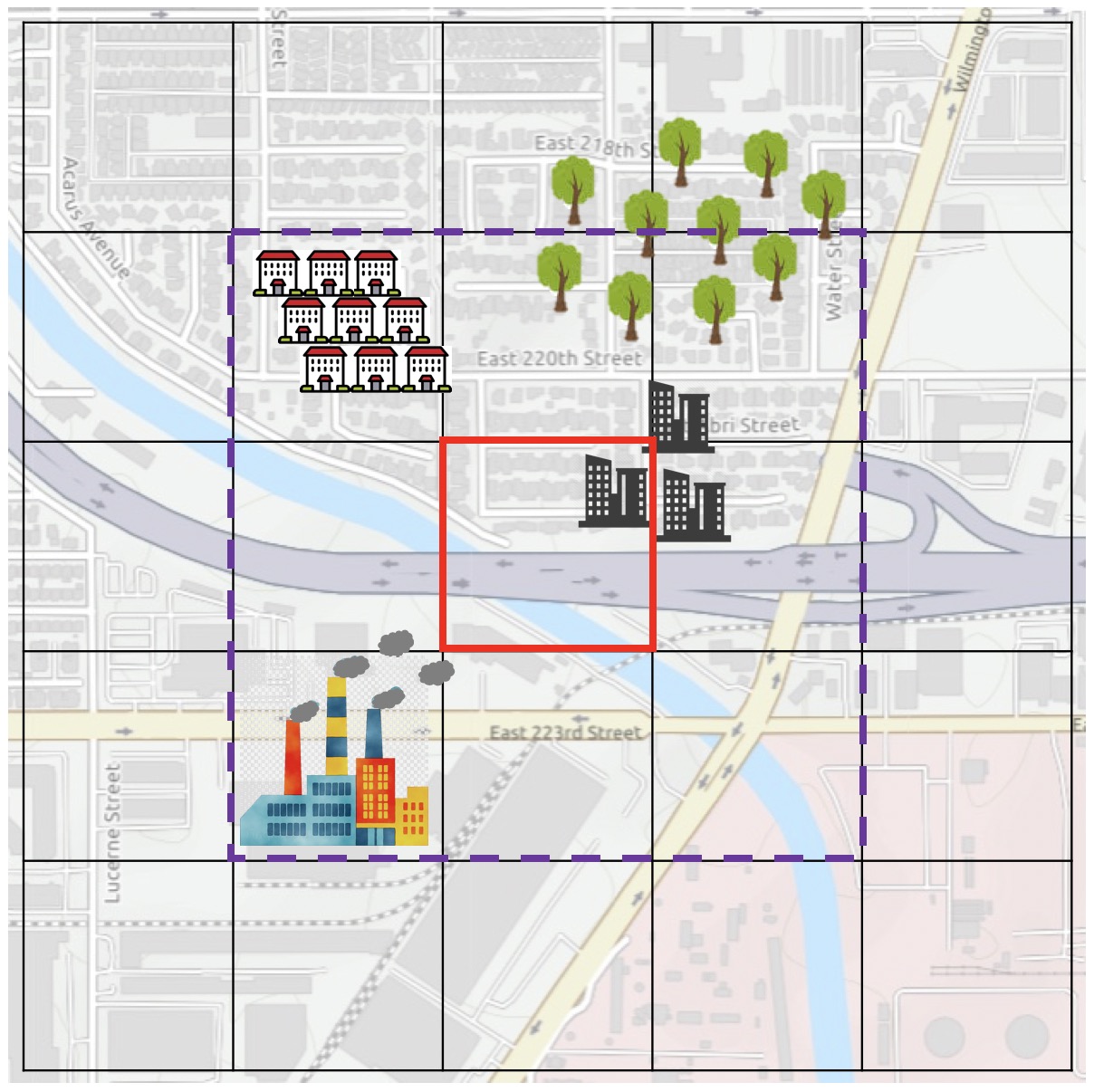}
            \vspace{-0.1in}            
            \label{figure: spatial_temporal_influence}
        \end{minipage}
    }
\vspace{-0.05in}
\caption{An example of learning spatiotemporal effects. The cell size is 500m$\times$500m. (a) A pollution emission from South-West power plant at time $T$. (b) The northeast cells are polluted at time $T+1$ with the northeast wind. (c) A process of learning environmental characteristics from one-step neighbors (within purple dotted box).}\label{figure: spatial_temporal_influence_all}
\vspace{-.15in}
\end{figure}

% \begin{figure}[ht]
%   \centering
%   \includegraphics[width=0.9\columnwidth]{figures/spatial_temporal_influence.jpg}
%   \caption{An example of learning spatial effects from one-step neighbors effects. The cell size is 500m$\times$500m. The red box is the target cell and the purple dotted box contains environmental characteristics of one-step neighbors.}
%   \label{figure: spatial_temporal_influence}  
% \end{figure}

% New content
\subsection{Autocorrelation-Guided Semi-Supervised Learning}\label{sec:ssl}
To handle the fact that the number of unlabeled cells is much greater than the labeled cells, typical semi-supervised learning methods could enforce nearby cells in space and time to have a similar air quality value~\cite{qi2018deep}. This is in line with the assumption that air quality measurements often have spatial autocorrelation that describes the tendency for areas or sites that are close together to have similar values.  Statistical methods, like Ordinary Kriging, take advantage of spatial autocorrelation to generate reliable estimates for the interpolated locations. However, it could be problematic when the observed locations are not evenly distributed or far apart from each other, or there is spatial non-stationarity in the data. 

In addition to incorporating local spatial autocorrelation, DeepLATTE proposes an autocorrelation-guided semi-supervised strategy that exploits the autocorrelation in the representation space of the spatiotemporal embeddings and enforces the predictions to embrace similar autocorrelation patterns to the observations in the representation space. The assumption is that the air quality values of two locations at the same or different times (or the same location at different times) could be similar if their environmental characteristics are similar (i.e., similar spatiotemporal embeddings) even when their spatial or temporal distances are large.
We define the autocorrelation as the relationship between the air quality variability and the similarity of spatiotemporal embeddings (the concept is close to the spatial autocorrelation, but our method does not directly use the physical distance). 
With this autocorrelation, we assume that similar spatiotemporal embeddings will generate close air quality values within some ``influence range'', and this pattern should be applicable for both observations and predictions. 

First, DeepLATTE decides the autocorrelation influence range by constructing a semivariogram to represent the autocorrelation in the real observations in the representation space of spatiotemporal embeddings. Figure~\ref{figure: semivariogram} shows an example semivariogram, of which the x-axis represents the distance (lag) between two spatiotemporal embeddings, $\boldsymbol{R}$. The distance lag is re-scaled to $[0, 1]$ and divided into multiple bins with a fixed lag size (e.g., 0.1). At a lag size, $h$, the y-axis represents the semivariance, $\gamma(h)$, for all the pairs of air quality values within that bin:
\begin{displaymath}
    \gamma(h) = \frac{1}{N(h)} \sum_{i} \sum_{j\ne i} {(Y(\boldsymbol{R}_i) - Y(\boldsymbol{R}_j))^2} 
\end{displaymath}

\noindent where $Y(\boldsymbol{R})$ is the air quality value (label) of the spatiotemporal embedding $\boldsymbol{R}$. The distance between $\boldsymbol{R}_i$ and $\boldsymbol{R}_j$ must be in the corresponding bin. $N(h)$ is the number of pairs of $\boldsymbol{R}$ in the bin. 

In Figure~\ref{figure: semivariogram}, the red points represent the semivariances at the binned distances. Then we estimate the relationship between semivariance and distance lag with a curve function to describe the autocorrelation process. The blue line is the estimated Gaussian process as an example:
\begin{displaymath}
    f(h)=n+s\times (1-\exp{(-\frac{h^2}{(r/2)^2})})
\end{displaymath}

\noindent where $h$ is the distance lag and $f(h)$ is the empirical semivariance. The parameter $n$ refers to the ``nugget'', which is the variability that cannot be explained by distance (by default $n=0$). The parameter $s$ refers to the ``sill'', representing the maximum observed variability. The parameter $r$ refers to the ``range'' at which the semivariance stops increasing. For example, the estimated influence range in Figure~\ref{figure: semivariogram} is 0.6, meaning that air quality values have strong autocorrelation at distances less than 0.6 while there is no or little autocorrelation beyond this influence range that can be ignored.

\begin{figure}[ht]
  \centering \vspace{-.1in}
  \includegraphics[width=0.9\columnwidth]{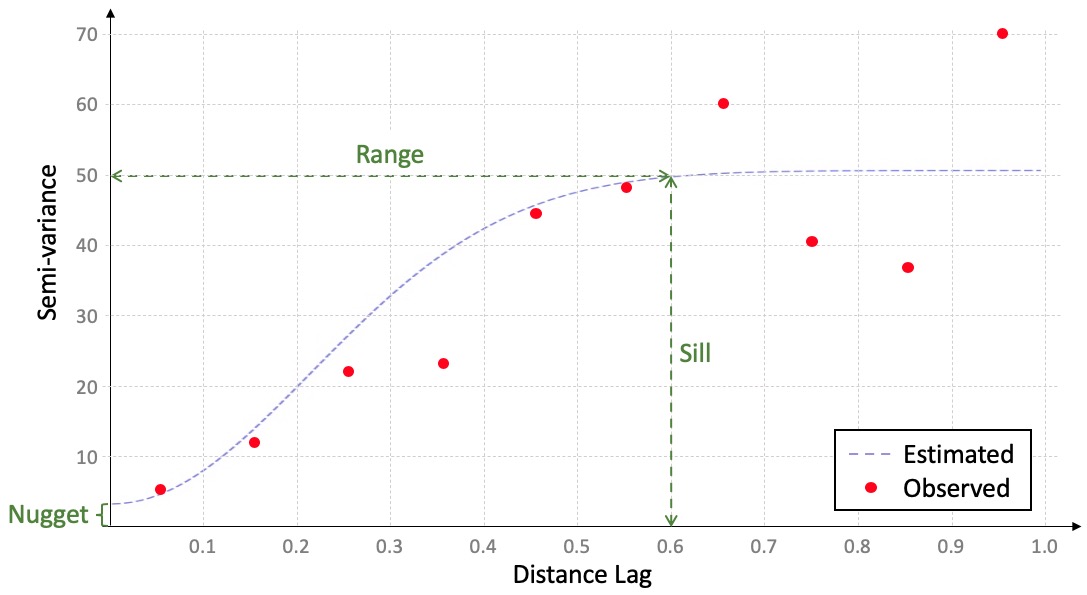}
  \vspace{-0.1in}
  \caption{An example semivariogram}
  \label{figure: semivariogram}  
  \vspace{-0.08in}
\end{figure}

After estimating the influence range from the semivariogram, DeepLATTE keeps the bins within the range and computes the distribution of semivariances for the labeled data and predictions separately in each of the kept bins. Specifically, DeepLATTE computes a Gaussian distribution $\mathcal{N}(\mu_{y_h}, \sigma_{y_h})$ to describe the distribution of the square errors between labels (i.e., $(y_i - y_j)^2$) when $ \left\| \boldsymbol{R}_i - \boldsymbol{R}_j \right\|_2$ is within distance lag $h$. Similarly, DeepLATTE computes $\mathcal{N}(\mu_{\hat{y}_h}, \sigma_{\hat{y}_h})$ for the predictions. 
The Gaussian distribution represents a generative model describing the autocorrelation pattern of the data at a given distance lag. Since the real observations and predictions should share similar autocorrelation patterns, DeepLATTE calculates the KL divergence to measure the distance between the two Gaussian distributions at each bin to quantify the difference between the autocorrelation of labels and predictions. Then DeepLATTE takes the summation of the KL terms from all the valid bins as the autocorrelation loss:
\vspace{-0.1in}

\begin{equation}
\begin{split}
\mathcal{L}_{ac} &= \sum_{h} D_{\mathrm{KL}} (\mathcal{N}(\mu_{y_h}, \sigma_{y_h}) || \mathcal{N}(\mu_{\hat{y}_h}, \sigma_{\hat{y}_h})) \\
&= \frac{1}{2} \sum_{h}(log\frac{\sigma_{\hat{y}_h}^2}{\sigma_{y_h}^2} - 1 + \frac{\sigma_{y_h}^2 + (\mu_{y_h}-\mu_{\hat{y}_h})^2}{\sigma_{\hat{y}_h}^2}) 
\end{split}
\label{eq: KL}
\end{equation}

\noindent where $\mu_{y_h}$ and $\sigma_{y_h}$ are mean and std. for the labeled pairs in lag distance $h$. $\mu_{\hat{y}_h}$ and $\sigma_{\hat{y}_h}$ are for the prediction pairs in $h$.

\subsection{DeepLATTE Overall Architecture}\label{sec:dl} 
The proposed network computes the supervised loss on labeled cells using the following cost function: 
\begin{equation}
    \mathcal{L}_{pred} = {\frac{1}{m}\sum_{i=1}^{m} (y_i-\hat{y}_i)^2}
\label{eq: train}
\end{equation}

\noindent where $m$ is the number of labeled cells, $\hat{y}$ is the prediction value of a cell, and $y$ is the real observation of the cell. 

The overall cost function of the proposed architecture is the sum of equation (\ref{eq: train}), (\ref{eq: sp}), (\ref{eq: ae}), (\ref{eq: stc}), and (\ref{eq: KL}): 
\begin{equation}
\mathcal{L} = \mathcal{L}_{pred} + \alpha \times \mathcal{L}_{sp} + \beta \times \mathcal{L}_{ae} + \lambda \times \mathcal{L}_{stc} + \eta \times \mathcal{L}_{ac}
\label{eq: overall}
\end{equation}

\noindent where $\mathcal{L}_{pred}$ is the loss over the training examples, $\mathcal{L}_{sp}$ is the loss from the sparse layer, $\mathcal{L}_{ae}$ is the reconstruction loss of the auto-encoder, $\mathcal{L}_{stc}$ is the loss from the representation constraint, and $\mathcal{L}_{ac}$ is the loss from applying the autocorrelation constraint; $\alpha$, $\beta$, $\lambda$, and $\eta$ are the hyper-parameters. We train the model by updating the network parameters using eq.~(\ref{eq: overall}).

\section{Experiments}
We implemented and tested DeepLATTE for air quality prediction in Los Angeles. \footnote{\url{https://github.com/spatial-computing/deeplatte-fine-scale-prediction}} In addition to its complex physical environment, Los Angeles is a well-studied area in terms of air pollution so there is an abundant literature that can be used to verify the prediction results (e.g., temporal and spatial distributions and trends) even when ground observations are limited.

%with Python 3.6 and the Pytorch framework. We performed the spatial computing processes in PostGIS (e.g., gridding and spatial aggregation).
\subsection{Experimental Setup}
\textbf{Datasets.} We conducted the experiments with the following open-source datasets covering a 50km$\times$40km region in Los Angeles (Figure~\ref{figure: testing_area}). (1) Air quality data: Our system collected hourly outdoor PM$_{2.5}$ concentrations in 2018 and 2019 from PurpleAir sensors.\footnote{\url{https://www2.purpleair.com/}} The available sensors increased over time, with the numbers ranging from 35 to 206. 
The PM$_{2.5}$ variations were generally higher during the winter (Nov., Dec., and Jan.) than the summer period (May, Jun., and Jul.).
(2) Meteorological data: Our system collected hourly weather data in 2018 and 2019 from DarkSky.\footnote{\url{https://darksky.net/dev}} DarkSky reports worldwide fine-scale weather data on various features, of which we extracted 10 features (e.g., temperature, humidity, and wind speed). The resolution of the requested weather data was of 5km$\times$5km, and our system cubically interpolated them to the target resolution.
(3) Geographic data: Our system extracted geographic information from OpenStreetMap to describe land uses, roads, traffic, railways, and water areas in various spatial representations (polygons, lines, and points). Each geographic feature contains various sub-types; for example, the ``roads'' feature type contains sub-types such as motorways, primary roads, and residential roads. We included a total of 80 geographic features.
(4) Other data: Our system included time information (i.e., hour of a day, day of a week, and day of a year) and the geo-coordinates of the grid center (i.e., longitude and latitude) for each combination of time \& location for the prediction task. 
% \vspace{0.01in}

% The input feature vector consists of a total of 14 dynamic features (10 meteorological features and 4 time features) and 82 static features (80 geographic features and 2 topological features). 
% \begin{figure}[h]
%   \centering
%   \includegraphics[width=0.95\columnwidth]{figures/monthly_mean_std.jpg}
%   \caption{2018-2019}
%   \label{Figure: monthly_mean_std}  
% \end{figure}

\textbf{Training Settings.}
Our system created a grid over the target region with a 500m$\times$500m cell size (total 6,992 cells). The sensor locations were unevenly distributed over the target region. To maintain the original spatial distribution of the sensors in both training and testing data, we divided the target region evenly into four areas (Figure~\ref{figure: testing_area_abcd}), from which we randomly picked 60\% of the available locations as training, 20\% as validation, and 20\% as testing. 

% For each test month, we trained a predictive model using the target-month sensor observations and contextual data. Also, if historical data from previous months were available, we first trained a model using data from the previous month and then fine-tune the model using data of the test month. For example, to predict PM$_{2.5}$ concentrations in Feb. 2018, we fine-tuned the model of Jan. 2018 with Feb. data.
For each month, we trained a predictive model using target-month sensor observations and contextual data. Also, if historical data from previous months were available, we leveraged the model trained from the previous months and fine-tuned the model using the target-month data. For example, to predict PM$_{2.5}$ concentrations in Feb. 2018, we fine-tuned the model of Jan. 2018 with Feb. data. We ensured that training and testing locations were mutually exclusive over these months.

% In the three-month setting, we trained the model with training observations in Jan., Feb., and Mar. and predicted for these months. By training with various amount of data, the goal is to examine if incorporating more environmental information from other time period would help improve model performance.

% To evaluate our prediction model on multiple spatial resolutions, we created two separate gridded surfaces over the same region with cell sizes of 500m$\times$500m and 1,000m$\times$1,000m. Table~\ref{Table: Details of Datasets} shows the details of the air quality datasets after mapping air quality sensor data to the grid maps.

To predict PM$_{2.5}$ concentrations at time $t$, we constructed the input features with temporal lags of six hours (i.e., from $t-6$ to $t$). We set the latent embedding of the auto-encoder as 32 neurons. We constructed each ConvLSTM containing one layer with 64 hidden states. We applied three kernel sizes (1$\times$1, 3$\times$3, and 5$\times$5), i.e., at most two-step neighbors (distance range is 1,500m). Therefore, the output size of the ConvLSTM component is 192. We set the threshold of the sparse layer as 0.0001, $K_T$ and $K_S$ as 1, and $\lambda_1=\lambda_2$. The initial learning rate 
was 0.001, with early stopping on the validation dataset. The hyper-parameters are chosen by grid search on the validation set and we reported the results with the set of hyper-parameters yielding the best average performance, that is $\alpha=1$, $\beta=5$, $\lambda=5$, and $\eta=0.1$. 

\begin{figure}[htbp]
\centering
    \subfigure[Target area]{
        \begin{minipage}[t]{0.46\columnwidth}
            \centering
            \includegraphics[width=\columnwidth]{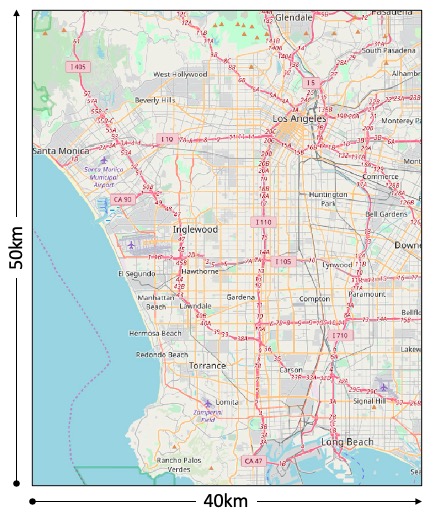}
            \vspace{-.15in}
            \label{figure: testing_area}
        \end{minipage}
    }
    \subfigure[Four sub-areas]{
        \begin{minipage}[t]{0.46\columnwidth}
            \centering
            \includegraphics[width=\columnwidth]{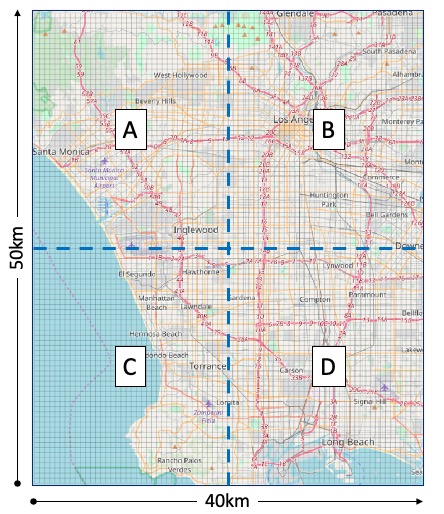}
            \vspace{-.15in}
            \label{figure: testing_area_abcd}
        \end{minipage}
    }
\vspace{-.08in}
\caption{(a) shows the prediction region in Los Angeles; (b) shows the equally-divided area as a gridded surface.}
\vspace{-0.1in}
\end{figure}

%%%%%%%%%%%%%%%%%%%%%%%%%%%%%%%%%%%%%%%%%%%%%%%%%%%%%%%%%%%%%%%%%%%%%%%%%%%%%%%%%
\subsection{Baseline Methods} 
We compared DeepLATTE with several state-of-the-art and commonly used models as well as DeepLATTE variants:

\textbf{Random Forest (RF).} \citet{yu2016raq} demonstrated that RF outperformed other approaches, including logistic regression, decision tree, and artificial neural network for fine-spatial-scale air quality prediction. % Therefore, RF is selected as one of the baseline methods. 

\textbf{Inverse Distance Weighting (IDW).} IDW is a standard spatial interpolation method that calculates predictions as the weighted average of the available measured points based on their spatial distance. We set the power parameter as 2.
% ${L}$ at time ${T}$ with the following equation:
% \begin{displaymath}
% Pred_{L, T} = {\frac{\sum_{i=1}^{n}{s_i \times w_i}}{\sum_{i=1}^{n}{w_{i}}}}
% \end{displaymath}
% where ${n}$ is the number of sensors, ${s_i}$ is the PM$_{2.5}$ value for the ${i^{th}}$ sensor at time ${T}$, and ${w_i}$ is the inverse spatial distance between the target location and the ${i^{th}}$ sensor.

\textbf{Ordinary Kriging (OK).} OK is a spatial statistical interpolation method that models the interpolated surface using a spatial process governed by the spatial autocorrelation of the measured points. We set the kernel function as Gaussian process and the number of bins as 10.

% \textbf{U-Air \cite{zheng2013u}.} A co-training based approach that combine spatial module and temporal module. 
% \textbf{ConvLSTM \cite{xingjian2015convolutional}.} A deep neural network that captures spatial features by convolution operations inside LSTM cells, which is widely used in spatiotemporal prediction.

\textbf{\citet{lin2017mining}.} This is a machine learning method that mines important PM$_{2.5}$-related geographic features as ``geo-context'' for fine-grained air quality prediction. We tested the approach with our datasets and picked the number of clusters with the Elbow method.

\textbf{Deep Air Learning (DAL) \cite{qi2018deep}.} DAL is a semi-supervised learning approach that leverages the information from the unlabeled data with spatiotemporal smoothing for the prediction. We re-implemented the approach and tested with our datasets. We did a similar way for hyper-parameter tuning and ended up with the set $\lambda_1=1$, $\alpha=10$, and $\beta=10$.

\textbf{DeepLATTE Variants.} (1) DeepLATTE without feature selection module; (2) DeepLATTE without the autocorrelation-guided constraint.

%%%%%%%%%%%%%%%%%%%%%%%%%%%%%%%%%%%%%%%%%%%%%%%%%%%%%%%%%%%%%%%%%%%%%%%%%%%%%%%%%
\subsection{Evaluation}
% For the cell locations where labeled data (real observations of air quality) are available, we used the following metrics to evaluate model performance:

\textbf{Quantitative Evaluation.} (1) We used the Root Mean Squared Error (RMSE) and R-Squared (R$^2$) scores to evaluate the model performance. 
% \noindent(1) Root Mean Squared Error (RMSE):
% \begin{displaymath}
% RMSE = \sqrt[]{\frac{\sum_{i=1}^{n}|y_{i} - \hat{y}_{i}|}{n}}
% \end{displaymath}
% \noindent(2) R-Squared (R$^2$) provides a measurement of how well the model fits the testing data or how much variability in PM$_{2.5}$ can be explained by the model.
% % Basically, it looks at the correlation between the outcome of the model and the ground truth. 
% We compute R$^2$ using \textit{R$^2$-score} in Python scikit-learn.
(2) We quantified the relationship between the variation in the fine-scale PM$_{2.5}$ predictions and the distance to critical environmental features, e.g., comparing the average predictions near highways within 500m and 1,000m, and evaluated the results with findings in the literature. 
% (3) We computed spatial autocorrelation in the areas with dense sensors to see if predictions preserve the spatial distribution in a small region as the sensor observations.

% \begin{displaymath}
% RMSE = \sqrt[]{\frac{\sum_{i=1}^{n}(y_{i} - \hat{y}_{i})^2}{n}}
% \end{displaymath}

% (2) R-Squared (R$^2$) provides a measurement of how well the model fits the testing data or how much variability in PM$_{2.5}$ can be explained by the model. We computed the R$^2$ score using \textit{R2-score} in Python scikit-learn.

\textbf{Qualitative Evaluation.} (1) We presented both spatial and temporal visualizations of the prediction results for verifying fine variations and trends in spatial and temporal neighborhoods using studies in the literature. (2) We verified the selected features with existing using studies in the literature.

\subsection{Results and Discussion} 
\textbf{Model Performance.}
Figure \ref{figure: rmse} and \ref{figure: r2} show the RMSE and R$^2$ for the monthly models of DeepLATTE and baseline methods in 2018 and 2019. We observe that RF has poor performance because it cannot handle complex relationships between the predictors adequately. OK outperforms IDW since OK specifically models the spatial autocorrelation instead of only relying on the spatial distance to compute the weight of measured locations. However, OK does not consider environmental characteristics and generates smoothing prediction surfaces (see Section Visualizations). DAL has competitive performance but is worse than DeepLATTE because DAL only enforces spatially and temporally nearby predictions to be identical, while DeepLATTE considers the autocorrelation of the air quality values in the representation space so that two locations or times having similar spatiotemporal embeddings can still have similar air quality values even when they are far away in space and time. \citet{lin2017mining} generate predictions in terms of environmental characteristics (geo-context), but the approach does not consider temporal dependencies. In general, DeepLATTE achieved the best performance compared to other baseline methods. 

% There are few failure cases that DeepLATTE is slightly worse than OK  because there exists a strong spatial autocorrelation in the observations and OK takes the advantage of that. However, we will demonstrate later that DeepLATTE generates more convincible fine-spatial-scale predictions in terms of environmental characteristics.

We compare DeepLATTE to its variants by removing the feature selection module and autocorrelation-guided constraint, respectively. We observe that the RMSE increases 1.8-5.1\% without the feature selection module (i.e., $\alpha$ is 0). Also, the number of selected features decreases when $\alpha$ increases, but a large $\alpha$ ($>1$) leads to worse performance. The feature selection module aims to explicitly remove irrelevant features at the beginning of the network to largely reduce the noise from these features during training. We manually remove the irrelevant features to train the models without the feature selection layer, which show little difference in the performance from DeepLATTE. In comparison, if we only keep these irrelevant features for prediction, the model performs much worse than DeepLATTE. The selected features also offer insights to the PM$_{2.5}$ contributors for the prediction task. Moreover, we observe that the RMSE increases 4.1-8.3\% without autocorrelation-guided constraint (i.e., setting $\eta$ to be 0), indicating that the constraint truly guides the network to generate more reliable predictions that have similar autocorrelation to the labels in the representation space. The model achieves worse performance when $\eta$ is \{0.01, 0.5\} than $\eta$ is 0.1. We further examine the similarity of the autocorrelation patterns between the labels and predictions with the semivariograms, which results in a high correlation of 0.73 for the range values and 0.85 for the sill values.

\begin{figure}[htbp]
\centering
    \subfigure[RMSE]{
    \begin{minipage}[t]{0.95\linewidth}
        \centering
        \includegraphics[width=\linewidth]{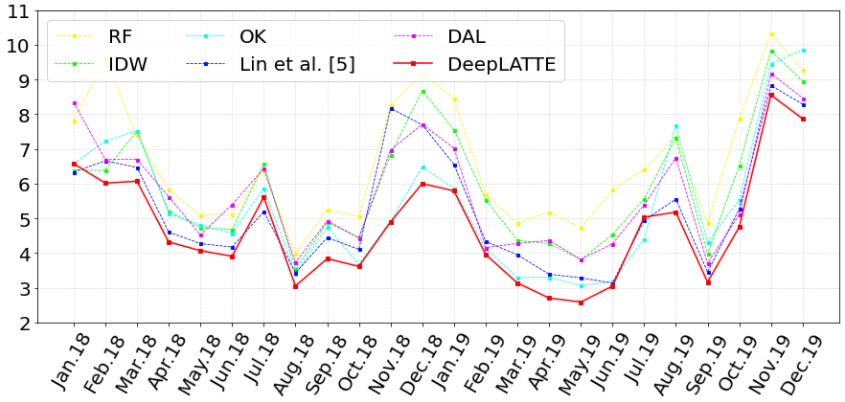}
        \vspace{-0.12in}
        \label{figure: rmse}
    \end{minipage}
    }
    \subfigure[R$^2$]{
        \begin{minipage}[t]{0.95\columnwidth}
            \centering
            \includegraphics[width=\linewidth]{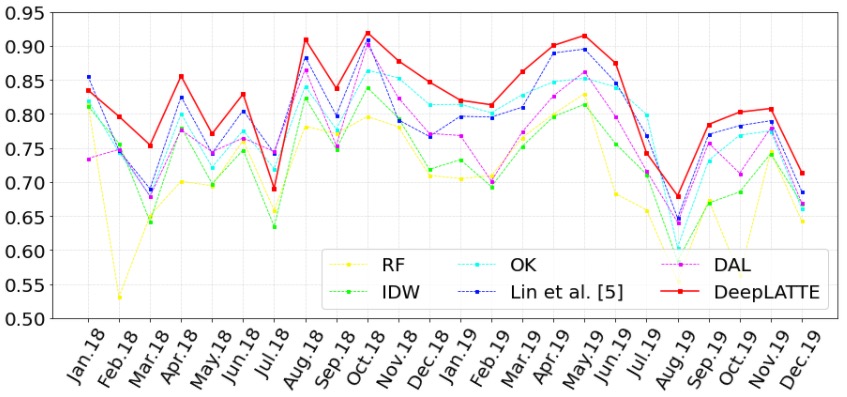}
            \vspace{-0.1in}
            \label{figure: r2}
        \end{minipage}
    }
\vspace{-0.1in}
\caption{Performance of DeepLATTE and baseline methods in RMSE and R$^2$}
\vspace{-0.12in}
\end{figure}

% Table~\ref{Table: RMSE Results} shows the comparison between DeepLATTE models using training data of various lengths. More specifically, we trained a model with three-month data and compared it to the combined results from one-month models (testing on the same locations over the target months). We observe that one-month models slightly outperform three-month models with 2\% improvement and six-month models with 4\%-6\% improvement. Therefore, incorporating more information (from previous time period) to generate one general model might smooth out the predictions and lead to lower performance but still with an acceptable accuracy. This would be a trade-off between a more accurate model and a more general model.

% \begin{table}[h]
% \centering
% \caption{.}
% \label{Table: RMSE Results}
% \renewcommand\arraystretch{1.1}
% \begin{tabular}{| l | c | c | c | c |}
%     \hline & \multicolumn{2}{c|}{Three-Month} & \multicolumn{2}{c|}{One-Month} \\ \hline
%      & RMSE & R2 & RMSE & R2 \\ \hline
%     % Jan.-Mar. 2018 &
%     % Apr.-Jun. 2018 &
%     Jul.-Sep. 2018 & 3.9798 & 0.8665 & \textbf{3.5502} & \textbf{0.8857} \\ \hline
%     Oct.-Dec. 2018 & 5.4360 & 0.8415 & \textbf{5.0957} & \textbf{0.8581} \\ \hline
%     & \multicolumn{2}{c|}{Six-Month} & \multicolumn{2}{c|}{One-Month} \\ \hline
%     Jan.-Jun. 2018 & 5.6798 & 0.7565 & \textbf{5.5502} & \textbf{0.7857} \\ \hline
%     Jul.-Dec. 2018 & 5.9183 & 0.8192 & \textbf{4.9976} & \textbf{0.8716} \\ \hline
%     \end{tabular}
% \end{table}
 
\textbf{Variations in Predictions.} Evaluating the model performance with limited observed locations is not enough to demonstrate the prediction ability. Since PM$_{2.5}$ concentrations are highly correlated with environmental characteristics~\cite{lin2017mining}, we examine the variations in predictions with varying distances to the geographic features (e.g., the blue areas in Figure~\ref{figure: corr_pm_feature}). Table~\ref{table: corr_pm_feature} shows the average predictions from DeepLATTE and OK (in Jan. 2019) with various distances to motorways, light rail, and parks. We observe that the predictions from DeepLATTE show high average values when the locations contain motorways and light rail and decrease when the locations are further from them. In contrast, the locations with parks have lower PM$_{2.5}$ predictions than others. Also, the predictions around parks are much better than motorways and light rail, which is consistent with the existing finding~\cite{nowak2010air}. Table~\ref{table: corr_pm_feature} shows that the predictions from OK have no apparent pattern regarding the environmental characteristics. Since OK generates smoothing surfaces of PM$_{2.5}$ concentrations, the average prediction values from OK are slightly larger than those from DeepLATTE. 

\begin{figure}[htbp]
  \centering
  \vspace{-0.05in}
  \includegraphics[width=0.8\columnwidth]{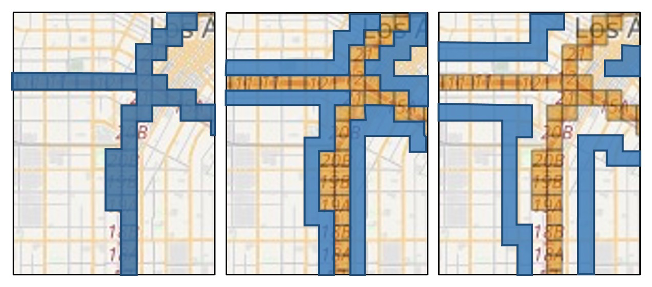}
  \vspace{-0.05in}
  \caption{An example of the locations containing motorways (left), one-step neighbors away from motorways (middle), and two-step away from motorways (right) in blue.}
  \vspace{-0.05in}
  \label{figure: corr_pm_feature}  
\end{figure}

\begin{table}[h]
\centering
\caption{Variations of predictions towards geographic features}
\label{table: corr_pm_feature}
\begin{tabular}{| c || c | c | c | c |}
\hline
    \multicolumn{2}{|c|}{} & \textbf{Motorways} & \textbf{Light Rail} & \textbf{Parks} \\ \hline
    \multirow{3}{*}{\rotatebox{90}{LATTE }} & zero-step & 17.03 & 17.39 & 14.87 \\ \cline{2-5}
    ~ & one-step & 16.65 & 16.55 & 15.79 \\ \cline{2-5}
    ~ & two-step & 16.44 & 16.92 & 15.54 \\ \hline \hline
    \multirow{3}{*}{\rotatebox{90}{OK}} & zero-step & 17.35 & 17.57 & 16.73 \\ \cline{2-5}
    ~ & one-step & 17.30 & 17.41 & 16.27 \\ \cline{2-5}
    ~ & two-step & 17.25 & 18.01 & 16.38 \\ \hline
\end{tabular}
\vspace{-0.1in}
\end{table}

% \textbf{Spatial Autocorrelation.} 
% We examine the spatial autocorrelation of our predictions verse the observed data in two small regions with relatively dense senors, i.e., Santa Monica and Downtown LA. 

\textbf{Visualizations.} Figures~\ref{figure: feb19} and~\ref{figure: oct19} spatially visualize the monthly average of the predictions from DeepLATTE while Figures~\ref{figures: ok_feb19} and~\ref{figures: ok_oct19} are the predictions from OK. We can see that the average predictions from OK are smooth over the region, which only offers a general idea about the variation of the PM$_{2.5}$ concentrations at a coarse spatial scale. 
In comparison, DeepLATTE predictions provide additional spatial details, consistent with some of the selected features. For example, higher PM$_{2.5}$ concentrations along freeways (I-110, I-10, I-5, and I-405) are observed due to traffic-related sources prevalent in Los Angeles~\cite{westerdahl2005mobile}. PM$_{2.5}$ values present an apparent pattern along railways (light rail). Also, the predictions show quite different patterns in February from October, i.e., Oct. has much worse air quality than Feb. in 2019, probably due to the different weather conditions in these two months. However, unlike the predictions from OK generating high values everywhere, DeepLATTE can predict low PM$_{2.5}$ concentrations in the north of the region because of the green land.
% By comparing the results between two resolutions, we can see that the grid maps show clear impacts of environmental characteristics on air quality when the resolution is higher, which might be because 1,000m$\times$1,000m cell size is too large for showing the spatial effects. 
In general, our model successfully generated predictions that present how the PM$_{2.5}$ concentrations vary with environmental characteristics over the space.

Figure~\ref{figure: feb_ts} shows the hourly average of PM$_{2.5}$ predictions in July and November over three small regions, i.e., Downtown LA (in Figure~\ref{figure: testing_area_abcd} sub-area~``B''), Santa Monica (in sub-area~``A''), and Long Beach (in sub-area~``D''). We observe that PM$_{2.5}$ concentrations generally reach the peak at 8 and 9 a.m. and minimum at 4 and 5 p.m.. This is because, in the morning, people start their activities (e.g., commuting) that produce PM$_{2.5}$, and in the afternoon, due to sea-breezes and mountain-induced flows, the wind takes the particle matter inland that leads to good air quality. Then the wind goes back from inland with accumulated particle matter that worsen air quality during the night~\cite{lu1995air}.
Besides, the air quality during the winter (Nov.) is much worse than the summer (Jul.), which is because winter has large temperature differences in a day that creates a thick layer in the air preventing PM$_{2.5}$ to escape~\cite{wallace2009effect}.
Also, Downtown LA and Long Beach have higher PM$_{2.5}$ concentrations than Santa Monica during the morning because of heavy traffic and industrial emission while Long Beach and Santa Monica are near the coastal areas, so PM$_{2.5}$ decreases with the wind patterns in the afternoon.

% time series comparison between our PM$_{2.5}$ predictions and the observed sensor concentrations at two testing locations in sub-area A and B (the location in A is near Santa Monica, and the location in B is near Downtown LA). We observe that DeepLATTE can capture different time-series patterns at locations that are far apart. Note that unlike methods as OK and \cite{lin2017mining} that build one model at a given time point, DeepLATTE is a general model that can generate predictions over the target period (one month in our experiment) at once. Our model performed well at most time points but inclined to underestimate when a sharp increase occurred because our data do not provide enough information for the model to infer the peak values. 

% For example, the observed peak between November 10$^{th}$ and November 11$^{th}$ was due to the Woolsey fire.\footnote{https://ktla.com/2018/11/11/unhealthy-air-plagues-much-of-southern-california-for-another-day-as-woolsey-fire-burns/} Both examples in Figure \ref{Figure: Time Series Prediction Results} show this peak, but our predicted concentrations are 10-20 $\mu$g/m$^3$ less than the sensor measurements. 

\begin{figure}[htbp]
\centering
    \subfigure[Feb. 2019, DeepLATTE]{
        \begin{minipage}[t]{0.46\columnwidth}
            \centering
            \includegraphics[width=\columnwidth]{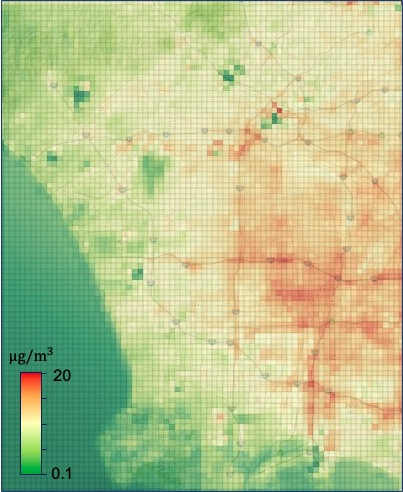}
            \vspace{-0.05in}
            \label{figure: feb19}
        \end{minipage}
    }
    \subfigure[Feb. 2019, Ordinary Kriging]{
        \begin{minipage}[t]{0.46\columnwidth}
            \centering
            \includegraphics[width=\columnwidth]{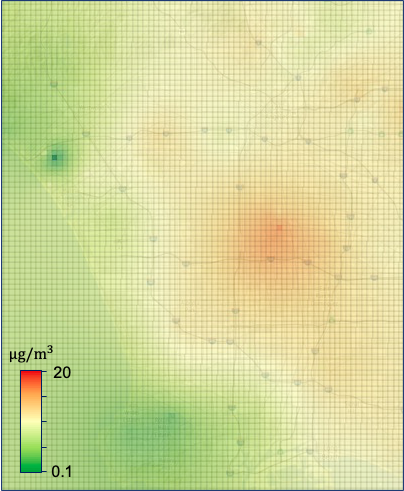}
            \vspace{-0.05in}
            \label{figures: ok_feb19}
        \end{minipage}
    }
    \vspace{.1in}
    \subfigure[Oct. 2019, DeepLATTE]{
        \begin{minipage}[t]{0.46\columnwidth}
            \centering
            \includegraphics[width=\columnwidth]{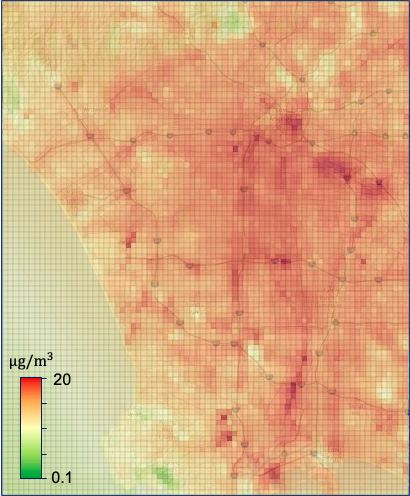}
            \vspace{-0.05in}
            \label{figure: oct19}
        \end{minipage}
    }
    \subfigure[Oct. 2019, Ordinary Kriging]{
        \begin{minipage}[t]{0.46\columnwidth}
            \centering
            \includegraphics[width=\columnwidth]{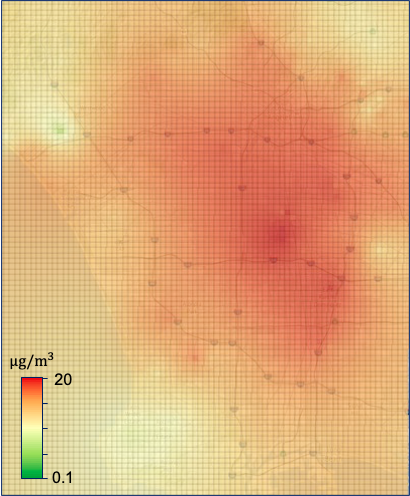}
            \vspace{-0.05in}
            \label{figures: ok_oct19}
        \end{minipage}
    }
\vspace{-0.12in}
\caption{Spatial visualizations of the predictions from DeepLATTE and OK}
\end{figure}

\begin{figure}[h]
  \vspace{-0.05in}
  \centering
  \includegraphics[width=0.93\columnwidth]{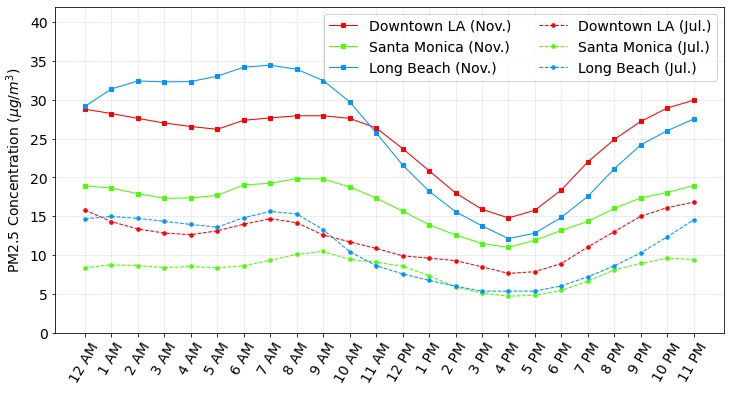}
  \vspace{-0.1in}
  \caption{Hourly pattern of PM$_{2.5}$ in Nov. and Jul. 2019 from DeepLATTE}
  \vspace{-0.1in}
  \label{figure: feb_ts}  
\end{figure}

\textbf{Selected Features.} 
Table~\ref{Table: Selected Features} shows the most frequent 10 selected dynamic features and 10 static features from all the DeepLATTE models ordered by the number of appearing times. The strong relationship between PM$_{2.5}$ concentrations and ``visibility'' is expected and has been seen previously~\cite{pui2014pm2}. \citet{megaritis2014linking} demonstrated PM$_{2.5}$ concentrations are highly related to ``temperature'', ``humidity'', ``pressure'', and ``wind speed'', which were automatically selected by DeepLATTE. Time information including weekday/weekend and day/nighttime is important since it indicates temporal variation and population mobility. 

In addition, most of the selected static features are consistent with existing studies on PM$_{2.5}$ in Los Angeles. Longitude and latitude are also important indicators since PM$_{2.5}$ concentrations are correlated with locations~\cite{zamani2019pm2}. DeepLATTE identified light rail is important for predicting PM$_{2.5}$, which has been demonstrated in \citet{kam2011particulate} that the light-rail lines are strongly associated with ambient PM levels in Los Angeles (R$^2$=0.61) by personally monitoring the air quality at the stations. Also, Moore et al.~\cite{moore2007land} showed that arterial roads and open green areas are statistically significantly associated with PM$_{2.5}$ in Los Angeles (R-value is approximately 0.4 to 0.6 respectively) using LUR approach. However, these existing studies require long-term and costly investigations by the environmental scientists with domain-specific expertise. In contrast, DeepLATTE automatically learns from a variety of raw features, selecting those that are most pertinent to air quality prediction. For other selected static features, such as ``traffic fuel'' (referring to gas stations) and ``traffic stop'' (referring to the intersections with stop signs), we do not find the existing work on analyzing the relationship between them and PM$_{2.5}$ concentrations to demonstrate our results, but the ``traffic fuel'' and ``traffic stop'' features intuitively relate to vehicle air pollution emissions. 

Furthermore, the selected features could vary each month. For example, during the summer time, DeepLATTE tends to select fewer features (around 20) for predicting PM$_{2.5}$ concentrations than the winter time (around 30). Visibility and humidity stand out when predicting for the summer period while time indicators are more likely to be selected for the winter period. This might be because the temporal variation of PM$_{2.5}$ in winter is higher than summer, the models require time information to generate reliable estimations.

\begin{table}[h]
\vspace{-0.05in}
\centering
\caption{Selected dynamic features}
\label{Table: Selected Features}
\begin{tabular}{|m{0.15\columnwidth} || m{0.7\columnwidth}|}
    \hline \textbf{Dynamic Features} & visibility, humidity, day of week, wind speed, hour of day, temperature, pressure, day of year, cloud cover, wind direction \\ \hline \hline
    \textbf{Static Features} & latitude, longitude, light rail, traffic fuel, motorway, traffic stop, primary road, secondary road, waterways river, park \\ \hline
  \end{tabular}
\vspace{-0.05in}
\end{table}

\section{Related Work}
\subsection{Spatiotemporal Modeling}
Deep learning approaches have shown promising results in modeling spatiotemporal relationships in data. Graph-based recurrent neural networks are often used to deal with spatiotemporal data in a graph representation and are flexible in modeling irregular graphs like road networks.~\citet{li2017dcrnn_traffic} propose a diffusion convolutional recurrent neural network (DCRNN) to forecast traffic volumes by combining the diffusion convolution operation with recurrent neural networks (RNN) on a spatial distance-based graph.~\citet{lin2018exploiting} extends DCRNN by constructing a geo-context-based graph to model the spatial relationship in the learned feature space for forecasting air quality values.~\citet{hsieh2015inferring} build an affinity graph by connecting nearby sensor measurements in space and time and apply a semi-supervised inference model to predict air quality values at unobserved locations. However, the network nature of graph-based approaches are not suitable for handling continuous surfaces (as either input or output) and are not scalable for fine-spatial-scale predictions that could have thousands of prediction locations. 

In contrast, image processing methods, like the convolution operation, is more suitable for spatiotemporal data represented in continuous surfaces (i.e., grids). ConvLSTM~\cite{xingjian2015convolutional} embeds the convolution operation directly in the LSTM network to capture spatiotemporal information simultaneously for weather forecasting. Other work leverages the ConvLSTM network both for forecasting (e.g.,~\cite{guo2019air, wang2018deep} and for fine-scale prediction (e.g.,~\cite{le2020spatiotemporal}). 
In addition, attention-based approaches also capture the spatial relationships by learning the attention score for the observed locations to predict for the target locations~\cite{pan2020spatio, cheng2018neural}. However, these approaches usually require large amounts of evenly distributed labeled data to achieve good performance, especially for the prediction tasks.

% \cite{yao2018deep} propose a deep multi-view network to predict taxi demand based on CNN and LSTM. 

\subsection{Air Quality Prediction}
Traditional spatial interpolation methods, like IDW and Ordinary Kriging, do not explicitly include environmental characteristics and fail to generate fine-scale predictions~\citep{lin2017mining}. Classical dispersion models, such as Gaussian Plume models~\citep{leelHossy2014dispersion}, require expert knowledge and resources that are not available everywhere, such as traffic data. Land-use regression models also rely on expert-selected predictors, which cannot be easily generalized to other geographic areas. In contrast, our data-driven approach can automatically mine air quality-related predictors in contextual data from multiple publicly available datasets.
In addition, machine learning approaches are widely used for air quality prediction. Lin et al.~\cite{lin2017mining} propose a method to identify significant air quality-related factors and apply them to generate predictions. However, the proposed approach does not take temporal effects into account.~\citet{zheng2013u} propose a co-training framework with separate classifiers for spatial and temporal features, which does not jointly model the spatiotemporal effects on air quality.~\citet{qi2018deep} propose a semi-supervised learning method with spatiotemporal smoothing. However, unlike DeepLATTE, this proposed method ignores the potential spatial autocorrelation patterns in air quality data and does not include predictors that describe the environment. 

\section{Conclusion and Future Work}
% This paper presented a novel approach, DeepLATTE, for fine-spatial-scale prediction of location-dependent time series data and an application of air quality prediction. The advantages of the presented architecture and application are that it: (1) takes advantage of autocorrelation in the air quality values in the representation space to guide the predictions to approach the pattern learned from labeled data in a semi-supervised way
% % , (2) automatically selects air quality-related factors from raw contextual data to help interpret the results, 
% and (2) considers spatial and temporal effects simultaneously and generates reliable spatiotemporal representations for prediction. The fine-scale predictions of PM$_{2.5}$ concentrations of DeepLATTE can be used in analyzing air quality-related health effects such as respiratory disease and asthma. Overall, the presented network architecture is flexible and could apply to many scientific prediction problems dealing with spatiotemporal phenomena with sparse ground truth measurements. 
% We plan to apply this approach to other data types such as to predict traffic and noise. We will also explore incorporating other sources of publicly-available contextual data including remote sensing observations.

This paper presented a novel approach, DeepLATTE, for fine-spatial-scale prediction of location-dependent time series data and an application of air quality prediction. The main contribution of the presented architecture is a novel method that learns a spatiotemporal representation from contextual data and labeled data and then uses the autocorrelation pattern in the representation space to guide the predictions in a semi-supervised way. This method allows for the integration of well-established spatial statistics tools with neural networks and enables accurate prediction of location-dependent time series data.  Overall, the presented network architecture is flexible and could apply to many scientific prediction problems dealing with spatiotemporal phenomena with sparse ground truth measurements. 
We plan to test this approach to other spatiotemporal phenomena, such as noise prediction. 
% We will also incorporate other sources of publicly-available contextual data, such as remote sensing observations. 

\section{Acknowledgement}
This work is supported by the NIH grant 1U24EB021996- 01 and Nvidia Corporation.

\bibliographystyle{IEEEtranN}
\bibliography{ref}

\end{document}